\pdfoutput=1

\documentclass[11pt]{article}

\usepackage[final]{acl2021}

\usepackage{times}
\usepackage{latexsym}

\usepackage[T1]{fontenc}

\usepackage[utf8]{inputenc}

\usepackage{microtype}

\usepackage{todonotes}

\usepackage{enumitem}
\usepackage{booktabs}
\usepackage{multirow}
\usepackage{amsmath}
\usepackage{amsfonts}
\usepackage{fontawesome}
\usepackage[ruled,vlined]{algorithm2e}
\usepackage{arydshln}
\usepackage[procnames]{listings}
\usepackage{multicol}
\usepackage[title]{appendix}
\usepackage{makecell}
\usepackage{color, colortbl}
\definecolor{Gray}{gray}{0.9}

\usepackage{longtable}
\usepackage{todonotes}

\usepackage{adjustbox}

\title{Intermediate Entity-based Sparse Interpretable Representation Learning}
\author{
Diego Garcia-Olano$^{1,3}\thanks{* Work completed during PhD at UT Austin}\,\,\,$
 Yasumasa Onoe$^1$ \\
\textbf{Joydeep Ghosh}$^1$ 
\textbf{Byron C. Wallace}$^2$ 
\\
$^1$University of Texas at Austin,
$^2$Northeastern University,
$^3$Meta AI\\
diegoolano@meta.com,
\{yasumasa,ghosh\}@utexas.edu,
b.wallace@northeastern.edu,
}

\date{}

\begin{document}
\maketitle
\begin{abstract}
Interpretable entity representations 
(IERs)  
are sparse embeddings that are ``human-readable'' in that dimensions correspond to fine-grained entity types and values are predicted probabilities that a given entity is of the corresponding type. 
These methods perform well in zero-shot and low supervision settings. Compared to  
standard dense neural embeddings, such interpretable representations may permit analysis and debugging. 
However, while fine-tuning sparse, interpretable representations improves accuracy on downstream tasks, it destroys the semantics of the dimensions which were enforced in pre-training. 
Can we maintain the interpretable semantics afforded by IERs while improving predictive performance on downstream tasks? Toward this end, we propose Intermediate enTity-based Sparse Interpretable Representation Learning (ItsIRL). 
ItsIRL realizes improved performance over prior IERs on biomedical tasks, while maintaining ``interpretability'' generally and their ability to support model debugging specifically.
The latter is enabled in part by the ability to perform ``counterfactual'' fine-grained entity type manipulation, which we explore in this work. 
Finally, we propose a method to construct entity type based class prototypes for revealing global semantic properties of classes learned by our model.\footnote{Code for pre-training and experiments available at \url{https://github.com/diegoolano/itsirl} }
\end{abstract}

\section{Introduction}


Deep pre-trained models yield SOTA performance on a range of NLP tasks, but do so by learning and exploiting dense continuous representations of inputs which complicate model interpretation. That is, the dimensions in learned representations have no \emph{a priori} semantics, and consequently are not directly human readable.
Indeed, this has inspired an entire line of work on ``probing'' dense representations to recover the implicit knowledge stored within them \cite{Fabio_Petroni_19, Nina_Poerner_19}. 

An alternative is to design architectures that explicitly imbue embeddings with semantics. 
To this end, recent work has proposed learning high-dimensional sparse interpretable entity representations (IERs) for general and biomedical domains \cite{onoe-durrett-2020-interpretable, garcia-olano-etal-2021-biomedical}.  
IERs are composed of a Transformer-based~\cite{Ashish_Vaswanir_17} entity typing model with a corresponding fine-grained static type system that accepts an entity mention and its context, and outputs individual probabilities that the mention is an instance of the respective types.
These embeddings may then be used as features for downstream tasks.

IERs afford a variety of model transparency (dimensions have semantics) which may facilitate model debugging and/or instill confidence in model outputs. 
For example, if one defines a linear layer on top of entity-type representations, learned coefficients are interpretable as weights assigned to specific entity types. 
One could learn rules or manually debug models by reviewing incorrect predictions and inspecting the corresponding induced representations to identify potentially systematic erroneous type assignments. 
In addition to providing this type of interpretability, IERs have been shown to perform comparatively well in zero- and few-shot settings \cite{onoe-durrett-2020-interpretable, garcia-olano-etal-2021-biomedical}.

A limitation of IERs is that they do not naturally permit fine-tuning, because doing so destroys the semantically meaningful entity typing representations learned during pre-training. 
This requirement is a limitation because fine-tuned models will in general achieve stronger predictive performance when supervision is available. 

In this work we aim to improve the predictive performance of IERs without sacrificing their interpretability. 
Specifically, we propose Intermediate enTity-based Sparse Interpretable Representation Learning (ItsIRL). 
We show that this model outperforms prior IERs by a substantial margin on experiments over biomedical datasets --- a domain where interpretability is often paramount --- while providing natural mechanisms for model debugging by virtue of the representational semantics inherent to the architecture. 

We then propose a counterfactual analysis of our intermediate interpretable layer to measure the effect of \emph{entity type manipulation} on downstream predictions.
This intervention is made possible by virtue of the model design. 
Using manually constructed, class-specific entity type sets we show that this intervention can be used to fix errors made by the proposed ItsIRL model automatically, ultimately allowing the model to outperform dense (uninterpretable) models in terms of test accuracy. 
We then propose a method in which we combine entity types over classes on training data to create positive and negative class prototypes that can be used to better understand 
the ``global'' semantics learned by ItsIRL for downstream tasks.

Our specific contributions are as follows: 
\begin{itemize}
    \item We introduce an intermediate interpretable layer into IERs; this layer output (representation) is then ``decoded'' into a dense layer which can be used for downstream predictions. 
    The decoding step can be fine-tuned for specific tasks. 
    \item We show that this approach empirically outperforms prior IER methods 
     on two diverse biomedical benchmark tasks, often by a substantial margin.
    \item We propose a counterfactual entity type manipulation analysis made possible by our architecture which 
    facilitates model debugging in an automated fashion with minimal, noisy supervision. This analysis allows our model to outperform dense (uninterpretable) models in terms of test accuracy and shows that the entity typing layer affects output classifications in an interpretable and intuitive way.
    \item We show how combining entity types over classes on the training set to create positive and negative class prototypes can be used to reveal task specific global semantics learned by our model.
\end{itemize}

\section{Background: Interpretable Entity Representations Model}

We first review the IER model architecture. Much of the material and notation here comes directly from \cite{onoe-durrett-2020-interpretable, garcia-olano-etal-2021-biomedical}. 
Let $s = (w_1, ..., w_N)$ denote a sequence of input context words, $m = (w_i, ..., w_j)$ denote an entity mention span in $s$ (over positions $i$ through $j$), and $\mathbf{t} \in  [0,1]^{|\mathcal{T}|}$ denote a vector whose values are predicted probabilities corresponding to fine-grained entity types $\mathcal{T}$ from a predefined type system.  

Given a labeled dataset $\mathcal{D} = \{(m, s, \mathbf{t}^*)^{(1)}, ... , (m, s, \mathbf{t}^*)^{(k)} \}$ the IERs' objective is to estimate parameters $\theta$ of a function $f_\theta$ that maps the mention $m$ and its context $s$ to a vector $\mathbf{t}$ that captures salient features (fine-grain types) of the entity mention within its context. The entity embedding $\mathbf{t}$ whose individual dimensions have explicit semantics can then be used directly as input for downstream tasks using standard similarity measures (e.g., dot products). 
Note that fine-tuning these representations would destroy their interpretability because dimensions would no longer be readable as the probability of the input representing specific entity types. 

\begin{figure*}[t]
    \centering
    \includegraphics[scale=0.6]{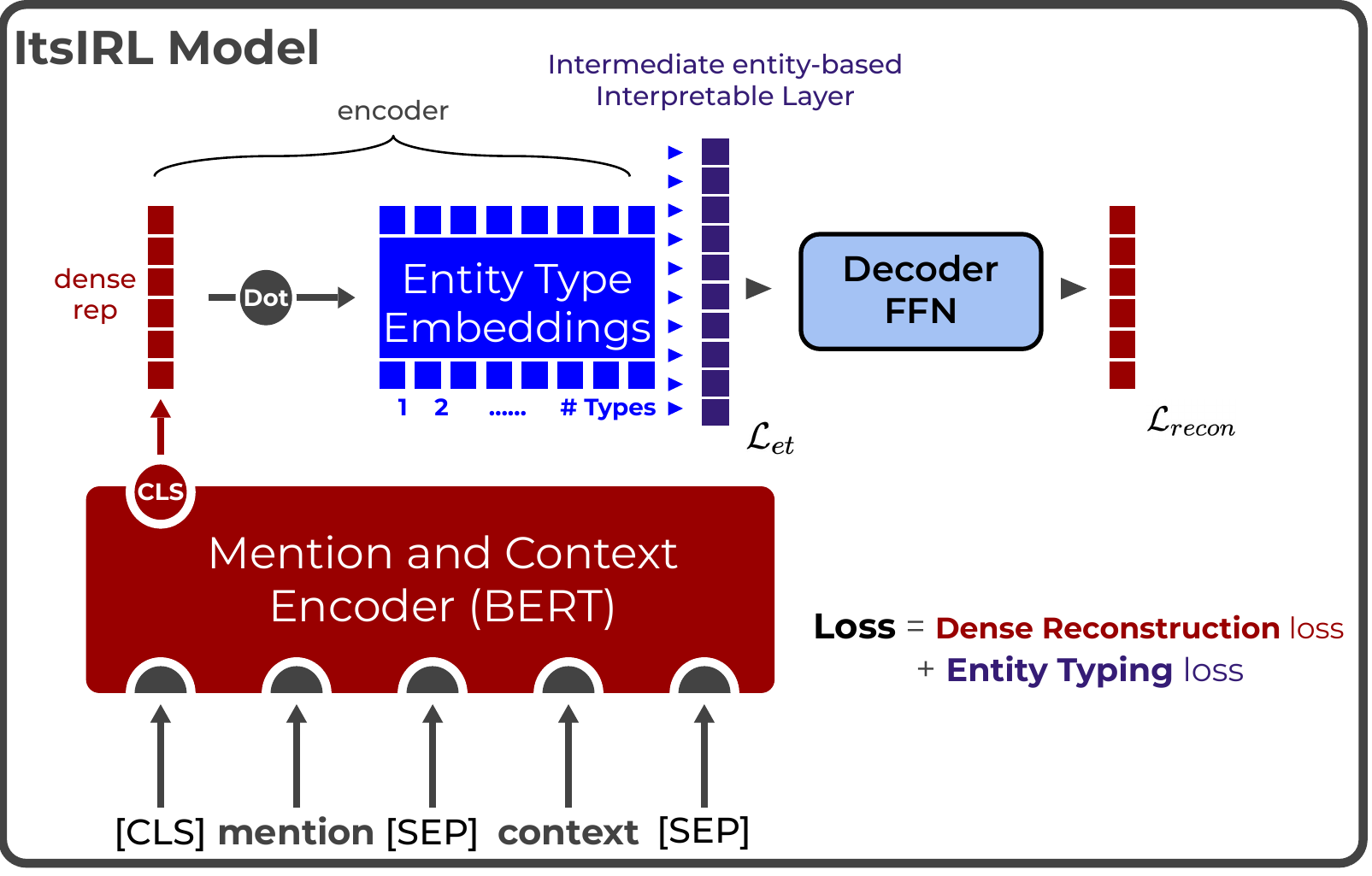}
        \caption{ItsIRL uses a LM and type supervision during pre-training to encode entity mention and context inputs for learning a matrix of entity type embeddings, an intermediate interpretable layer of type scores and a decoder to reconstruct the initial LM representation. The decoder can be fine-tuned on downstream tasks for better performance than IERs while keeping the semantics of the type layer.}
    \label{fig:sys_arch_v2}
\end{figure*}

The model $f_\theta$ that produces these embeddings is depicted as the ``encoder" in Figure~\ref{fig:sys_arch_v2}. 
First, a BERT-based encoder~\citep{ Jacob_Devlin_19} maps inputs $m$ and $s$ to an intermediate dense vector representation. 
The encoder input is a token sequence $\mathbf{x} =$ {\tt[CLS]} $m$ {\tt[SEP]} $s$ {\tt[SEP]}, where the mention $m$ and context $s$ are segmented into WordPiece tokens \citep{Yonghui_Wu_16}. 
The vector output {\tt [CLS]} token serves as a $d$-dimensional dense mention and context representation: $\mathbf{h}_{\texttt{[CLS]}} = \textsc{BertEncoder}(\mathbf{x}) \in \mathcal{R}^d$.  

The key ingredient of IERs is a \emph{type embedding layer}, which projects this intermediate representation to a vector whose dimensions correspond to the entity types in $\mathcal{T}$ using a single linear layer with parameters 
$\mathbf{E} \in  \mathcal{R}^{|\mathcal{T}| \times d}$.
Finally, each dimension (individually) is passed through the sigmoid function, yielding the 
predicted probabilities that form the interpretable entity representation $\mathbf{t}$ (the ``intermediate layer'' in Figure \ref{fig:sys_arch_v2}).
More concisely: 
$\mathbf{t} = \sigma \left(\mathbf{E} \cdot
\mathbf{h}_{\texttt{[CLS]}}\right)$. 
To estimate parameters we optimize the sum of binary cross-entropy losses entity types $\mathcal{T}$ over training examples $\mathcal{D}$.

\section{Intermediate Entity-based Sparse Interpretable Representation Learning}

We modify the IER model just described 
as follows: 
\begin{itemize}
    \item We project down the sparse entity typing layer and add pass its output through a three layer feed forward ``decoder'' network.
    \item We add an additional reconstruction component to our loss which is simply the mean squared error between the model's output and the initial {\tt [CLS]} representation given by the Transformer based model. 
\end{itemize}
This proposed model architecture --- which we have called ItsIRL --- is depicted in Figure \ref{fig:sys_arch_v2}.  
During pre-training, we adopt a loss $\mathcal{L}$ that combines entity typing loss over the sparse intermediate interpretable layer $\mathcal{L}_{\text{et}}$ and the reconstruction loss of the output representation $\mathcal{L}_{\text{recon}}$
 \begin{equation*}
 \mathcal{L} = \mathcal{L}_{\text{recon}} + \lambda  \mathcal{L}_{\text{et}}
 \end{equation*}
 where $\lambda$ is a hyperparameter to be tuned.
 
 The motivation behind the additional reconstruction loss is to pre-train a sort of auto-encoder with a sparse, high dimensional, interpretable latent space and rich dense output representations. 
 Here the encoder induces a sparse embedding of entity types as in prior work on IERs, but now for downstream tasks we can freeze the encoder (which yields  interpretable entity representations) and \emph{fine-tune the decoder}.
 That hope is that this allows for both interpretable entity types and improved task performance.  
 
 In contrast to prior IER work in which sparse entity type representations were used directly for downstream tasks, here we pass the intermediate interpretable representation into a feed forward decoder network that produces a new representation which is used for prediction. 
 This choice leads to differences in interpretability between IERs and our proposed architecture. 
 We explore this in Section \ref{counterfactualsection}, along with how these intermediate predicted entity types affect task performance and how user or automated mechanisms to manipulate (i.e., up or down weight) these intermediate types affects performance. 
 
 This approach in some ways resembles \emph{concept bottleneck} models (\citealt{conceptbottlenecks, conceptwhitening}; reviewed further in Section \ref{section:related-work}). However, these methods generally use low dimensional, \emph{human-labeled concept supervision} to guide learning \emph{for a single task}. By contrast, in our approach we exploit  large-scale, possibly noisy entity type supervision to learn to induce interpretable representations which might be useful across tasks, i.e., for general pre-training. 

We could pre-train such models in a few ways: (i) Train them end-to-end, or, (ii) Use existing IER models as points of initialization. 
In the latter case, we freeze the IER model originally trained using only $\mathcal{L}_{\text{et}}$ and train/update the rest of the model weights using only $\mathcal{L}_{\text{recon}}$ as the loss on our pre-training data.

For our experiments we use the publicly available biomedical IER model checkpoint, entity type system, and pre-training data from \cite{garcia-olano-etal-2021-biomedical}. 
The model checkpoint is based on an underlying PubMedBERT model \cite{gu2020domainspecific}. 
The type system contains 68,304 entity types and the training data consists of 37,357,141 triples of the form ({\tt mention}, {\tt context}, {\tt [list of entity types]}) derived from PubMed linked Wikipedia pages where entity types are Wikipedia categories.

\section{Experimental Setup}
We evaluate the proposed ItsIRL architecture on two biomedical benchmark tasks: Entity label classification for Cancer Genetics \cite{pyysalo-etal-2013-overview} and sentence similarity regression for the BIOSSES dataset found in the BLURB benchmark \cite{gu2020domainspecific}. 

\subsection{Cancer Genetics Entity Label Classification} 
The Cancer Genetics dataset \cite{pyysalo-etal-2013-overview} consists of 10,935 training, 3,634 dev, and 6,955 test examples from 300, 100, and 200 unique PubMed articles, respectively.
Given an article title/abstract and an entity mention, the objective is to categorize the entity into one of 16 classes which cover different subdomains in cancer biology.

For the downstream task we simply add a linear layer that accepts as its input the output of our pre-trained ItsIRL model and we then fine-tune the ItsIRL decoder and linear layer to minimize cross entropy loss.  
We stop training when the model accuracy ceases to improve on the dev set. 
We also provide numbers for how ItsIRL performs if we fine-tune on training data in an end-to-end fashion (ItsIRL E2E; i.e., unfreezing and updating the encoder weights and intermediate type layer); this destroys the interpretability of the intermediate layer enforced in pre-training. 
Results for using the prior Biomedical Interpretable Entity Representations (BIERs) dot product based model and PubMedBERT dense model are from \cite{garcia-olano-etal-2021-biomedical}. 
We provide ablations to 
explore the effect of decoder network layer size and pre-training.

\textbf{Results} We report task results in Table \ref{tab:hoc-res}. 
Compared to the prior IERs work (87.5\%), the ItsIRL model gives improved performance (91.9\%) while keeping the semantic interpretable entity type layer intact. 
ItsIRL E2E realizes performance comparable to fine-tuning PubMedBERT alone (95.7\% and 96.1\%, respectively), but in both cases we no longer have interpretable models which can be diagnosed and fixed at run time.  

As a point of reference, we also report results achieved by dense models. 
However, we emphasize that these do not provide the transparency afforded by ItsIRL; we are interested in achieving both accuracy \emph{and} interpretability --- models which strictly optimize the former may be viewed as a reasonable ``upper-bound'' with respect to accuracy alone, and in general we expect that realizing interpretability (and specifically in our case, ``debuggability'') will entail some trade-off in accuracy.\footnote{Related works (e.g., \citealt{conceptbottlenecks,senn2018} have tended to report results for \emph{only} other ``interpretable'' models as baselines; we include standard dense models here for completeness.}

We observe this expected trade-off here (ItsIRL performs better than BIER, but worse than end-to-end models which lack semantic representations). 
We also confirm that the proposed model can be fine-tuned end-to-end to achieve the same accuracy as the dense PubMedBERT model, at the expense of interpretability.
Perhaps more interestingly, in section \ref{counterfactualsection} we show that leveraging entity type manipulation at inference time allows the ItsIRL model to outperform both uninterpretable models.

We perform a few ablations to assess which parts of ItsIRL affect performance. 
We perform fine-tuning on the task data using a decoder whose weights are randomly initialized to test the effect of pre-training on 37 million triples.
The bottom of Table \ref{tab:hoc-res} shows that this degrades performance (88.9\% vs. 91.9\%) and suggests that pre-training the decoder network is important for task performance.

We additionally explored varying layer depths for our decoder (3, 5, 8) and observed similar performance across them; we therefore opted to use the smaller decoder network of 3 layers. 
We note that prior work \cite{garcia-olano-etal-2021-biomedical} 
explored adding a single linear layer on top of the entity type representation (which is identical to ours) and fine-tuning it for the task.  
This single layer ``decoder'' yields 68.1\% test accuracy, indicating that the additional network capacity and pre-training are both important.

\renewcommand{\arraystretch}{1}
\begin{table*}[t]
\parbox[t][][t]{.52\linewidth}{%
\small
\centering
\begin{tabular}{lcc} 
\toprule 
Model & \faSearch & Test Acc \\
\midrule
BIER-PMB* & \checkmark & 87.5  \\
ItsIRL & \checkmark & 91.9  \\
ItsIRL E2E* & - & 95.7 \\
PubMedBERT & - & 96.1 \\
\midrule
&\\
\midrule
Ablations & & Test Acc\\
\midrule
ItsIRL - random init & & 88.9\\
ItsIRL - 1 layer decoder&  & 68.1\\
\bottomrule
\end{tabular}
\captionsetup{justification=centering}
\caption{Cancer Genetics results\newline \faSearch\,= interpretable types }
\label{tab:hoc-res}
}
\hfill
\parbox[t][][s]{.52\linewidth}{%
\vspace{-62pt}
\small
\begin{tabular}{llllll} 
\toprule
\multicolumn{3}{l}{} & \multicolumn{3}{c}{Type Sparsity}\\
\cmidrule(r){4-6}
Model & \faSearch & MSE & @.01 & @.1 & @.25 \\
\midrule
BIER-PMB* & \checkmark & 5.05 & 33.6 & 8.1 & 4.4 \\
ItsIRL & \checkmark & 1.59 & 33.6 & 8.1 & 4.4 \\
ItsIRL E2E* & -  & 1.15 & 5723 & 780 & 330 \\
PubMedBERT & - & 1.14 & - & - & - \\
\bottomrule
\end{tabular}
\captionsetup{justification=centering}
\caption{BIOSSES sentence similarity results. \newline \newline PMB* = PubMedBERT\newline E2E* = End-To-End fine-tuned}
\label{tab:bio-res}
}
\end{table*}

\subsection{BIOSSES sentence similarity regression}
The Sentence Similarity Estimation System for the Biomedical Domain \cite{biosses} (BIOSSES) contains 100 pairs of PubMed sentences, each annotated by five expert annotators with an estimated similarity score in the range from 0 (no relation) to 4 (equivalent meanings). 
Predicting these scores (averaged over annotators) is a regression task used in the BLURB benchmark \cite{gu2020domainspecific}.

We use the train/dev/test splits from the BLUE benchmark \cite{peng-etal-2019-transfer}. 
We feed each sentence pair with a \texttt{SEP} between them as input and use mean squared error as our loss and for evaluation purposes amongst our model variants. 
In contrast to the Cancer Genetics task which has $>$10k training samples, 
this dataset is small, comprising 64, 16, and 20 train, dev, and test instances, respectively. 
We also evaluate the sparsity of the entity type layer induced by ItsIRL using different thresholds to numerically quantify the interpretability of these entity types, where having fewer types is more easily human interpretable.\footnote{As the prior BIER-PubMedBERT and ItsIRL share the same model checkpoint and hence interpretable entity typing layer, BIER-PMB will have the same type sparsity as ItsIRL.} 
Entity types whose weights are larger than a threshold are semantically meaningful at that threshold.

\paragraph{Results} We show results for the sentence similarity regression task in Table \ref{tab:bio-res}. The pattern in our results is similar to above: ItsIRL outperforms BIERs due to its being fine-tuned on task specific data. 
ItsIRL is competitive with, but slightly underperforms, 
the end-to-end fine-tuned ItsIRL E2E variant and the dense PubMedBERT model (neither of these offer an interpretable entity layer after fine-tuning). 

In Table \ref{tab:bio-res} we also observe that the number of entity types shown to be semantically meaningful is much less and hence more interpretable when comparing ItsIRL with ItsIRL E2E which removes the semantic meaning of the entity types space.  
Figure \ref{fig:sparsity} in the Appendix shows this sparsity value as a percentage over many different thresholds, showing the fine-tuned ItsIRL is more sparse and interpretable than both the ItsIRL E2E model and the dense non-interpretable PubMedBERT model. 

\section{Entity Type Counterfactual Manipulation and Global Explainability}
\label{counterfactualsection}

We have claimed that (sparse) entity type representations permit ``interpretability'', but this is an ill-defined term in general.
Here we demonstrate that ItsIRL provides a specific type of ``interpretability'' in that it can help facilitate model understanding and error analysis via ``counterfactual'' entity type manipulation, made possible by the intermediate entity type layer. 
Specifically, we consider the Cancer Genetics classification task \cite{pyysalo-etal-2013-overview}, and focus on 
revealing learned global structure of classes. 
We then show how manipulating predicted types on erroneous test cases affects the ItsIRL model's performance.


\subsection{Entity Type Global Explainability}

To better understand the representations learned by ItsIRL for each class, we apply the task, decoder fine-tuned model over the training data.  
We gather all correctly predicted instances for each class, sum their interpretable entity type representations and normalize them.\footnote{$ \text{Positive class prototype}=\frac{\text{v} - \text{min(v)}}{\text{max(v)} - \text{min(v)}}$ where v is the sum of entity type representations for correctly predicted training instances
of a given class.} We refer to each of these as a \emph{positive class prototype}.  

\paragraph{Results} In Table \ref{tab:pos8sm} we show the ``top" entity types --- those with the highest weights --- for 7 of 16 class prototypes (for space); on inspection, these intuitively seem semantically meaningful with respect to the classes. 
In Appendix Table \ref{tab:pos8full} we also show the weights and index of each entity type in the 68k type system, with lower indices denoting types that appeared more often in pre-training data. 
We also provide the F1 scores and support of these classes on the test set. 
Looking at the indices of top entity terms per class prototypes, we note that they tend to be in the tens or hundreds range, implying that more frequent entity types in the training data dominate the positive prototypes. 
However, consider two classes for which we observe lower than average F1 scores: \emph{Multi-tissue structure} and \emph{Tissue}.
These prototypes include rare ``top'' entity types (e.g., ``soft tissue'', ``nephron'' and ``barcode'') with indices in the 1000s (3067, 1951 \& 2351) that were seen less during pre-training which shows the model may have learned weaker representations for entity types 
that appeared less frequently.


\begin{table*}[]
\centering
\resizebox{\textwidth}{!}{%
\begin{tabular}{llllllll}
\toprule
&
\makecell[tl]{Gene or\\gene product} & Cell & Cancer & \makecell[tl]{Simple\\chemical} & Organism & \makecell[tl]{Multi-tissue\\structure} & Tissue  \\
\midrule
1 &
\makecell[tl]{protein} &              
\makecell[tl]{cell} &                     
\makecell[tl]{disease} &                  
\makecell[tl]{ingredient} &         
\makecell[tl]{taxonomy} &                      
\makecell[tl]{blood} &        
\makecell[tl]{tissue} \\
2 &
\makecell[tl]{ingredient} &     
\makecell[tl]{elementary\\particle} &                  \makecell[tl]{neoplasm} &                     
\makecell[tl]{acid} &  
\makecell[tl]{mammals\\in 1758} &               
\makecell[tl]{angiology} &          
\makecell[tl]{cell} \\
3 &
\makecell[tl]{human} &             
\makecell[tl]{human cells} &                 
\makecell[tl]{oncology} &                       
\makecell[tl]{rtt} &                  
\makecell[tl]{humans} &            
\makecell[tl]{soft tissue} &             
\makecell[tl]{human body} \\
4 &
\makecell[tl]{gene} &   
\makecell[tl]{battery} &         
\makecell[tl]{tissue} &  
\makecell[tl]{who essential\\medicines} &      
\makecell[tl]{tool-using\\mammals} &                
\makecell[tl]{nephron} &     
\makecell[tl]{connective\\tissue}\\
5 &
\makecell[tl]{coagulation} &                      
\makecell[tl]{gene} &   
\makecell[tl]{abnormality} &        
\makecell[tl]{chemical\\compound} &  
\makecell[tl]{anatomically\\modern humans} &            \makecell[tl]{blood\\vessel} &      
\makecell[tl]{endocrine\\system} \\
6 &
\makecell[tl]{cell} &                   
\makecell[tl]{protein} &                    
\makecell[tl]{cancer} &               
\makecell[tl]{measurement} &           
\makecell[tl]{postmodernism} &               
\makecell[tl]{human body} &            
\makecell[tl]{epithelium} \\
7 &
\makecell[tl]{cell growth} &                
\makecell[tl]{pancreas} &                 
\makecell[tl]{syndrome} &       
\makecell[tl]{calcium} &                  
\makecell[tl]{patient} &         
\makecell[tl]{lymphatic sys} &             
\makecell[tl]{angiology} \\
8 &
\makecell[tl]{endothelium} &                  
\makecell[tl]{system} &               
\makecell[tl]{malignancy} &                  
\makecell[tl]{hydroxyl} &       
\makecell[tl]{medical term.} &         
\makecell[tl]{lymphoid org.} &          
\makecell[tl]{blood vessel}  \\
9 &
\makecell[tl]{homology} &         
\makecell[tl]{carboxylic\\acid} &              
\makecell[tl]{cell\\growth} &                 
\makecell[tl]{glucose} &         
\makecell[tl]{prothrombin\\time} &  
\makecell[tl]{mononuclear\\phagocyte sys} &             \makecell[tl]{histology} \\
10 &
\makecell[tl]{oncogene} &                   
\makecell[tl]{ester} &  
\makecell[tl]{paraneoplastic\\ syndromes} &             
\makecell[tl]{methyl group} &                     
\makecell[tl]{bbc} &                   
\makecell[tl]{gland} &              
\makecell[tl]{barcode}  \\

\bottomrule
\end{tabular}
}

\caption{Top 10 Entity Types by weight for 7 most frequent positive Prototype class embeddings}
\label{tab:pos8sm}
\end{table*}

Similarly, we can gather all training predictions that were incorrect, group them by the true labels, and then sum and normalize their entity type layers to generate negative prototypes. In Appendix Table \ref{tab:neg7} we show the most common error patterns and their negative prototypes' most important entity types.  We note the negative prototypes predicted align with the positive prototypes true classes.  

\begin{figure}[!t]
    \includegraphics[width=1.0\linewidth]{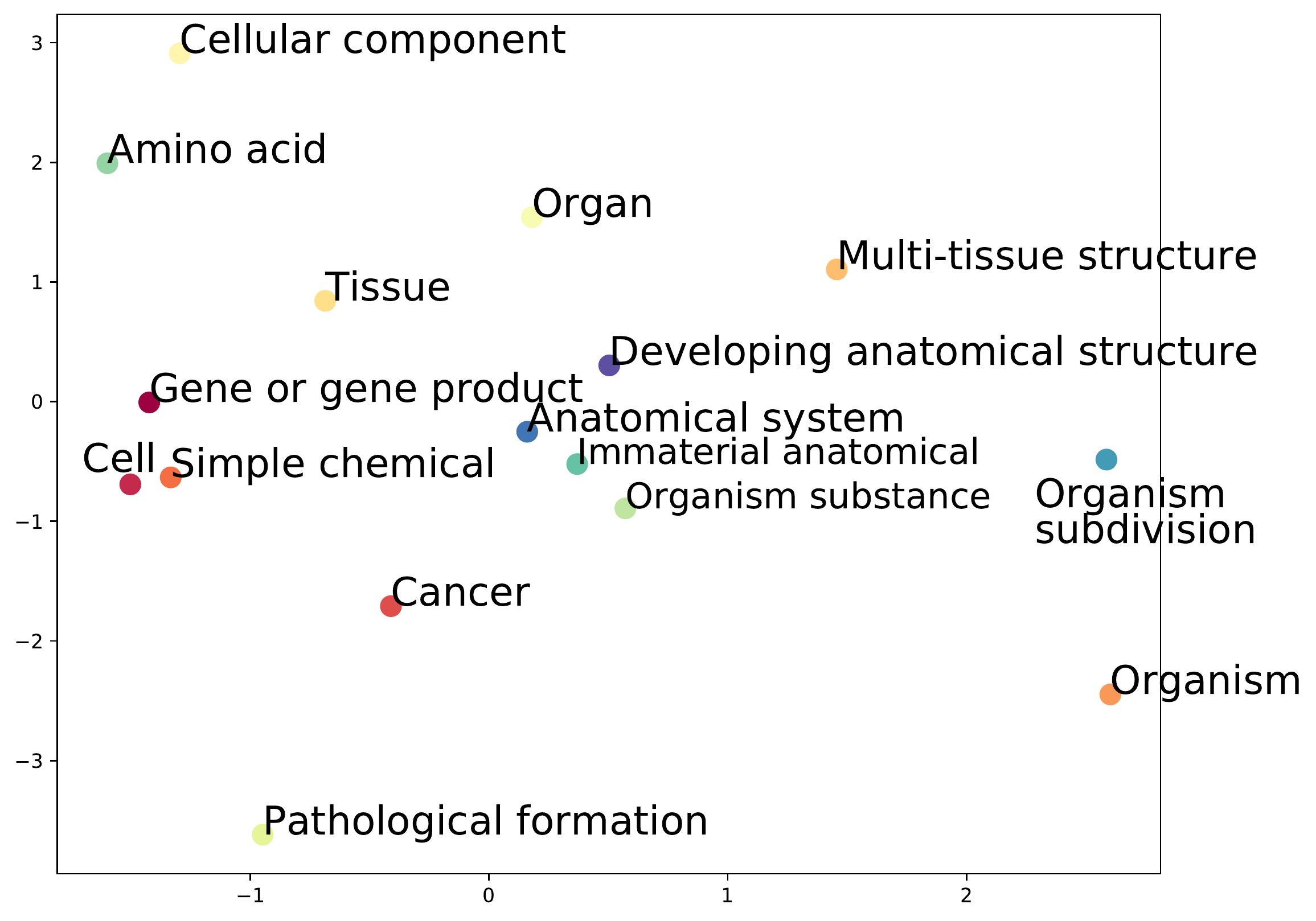}
        \caption{Positive Class Prototypes in 2D via PaCMAP}
    \label{fig:pacmap}
\end{figure}

Finally, in Figure \ref{fig:pacmap} we use PaCMAP \cite{pacmap} to visualize our positive prototypes in two dimensions.\footnote{PacMAP is a dimensionality reduction method shown to preserve the global and local structure of the data in its original space better than techniques such as TSNE and UMAP.}  
The distance between classes aligns well with the most common error patterns (i.e., Cell, Cancer, Chemical, and Gene cluster near each other) while ``anatomical'' and ``organism'' related classes also cluster near each other.

\subsection{Counterfactual Entity Type Manipulation} To explore how intermediate entity types affect downstream performance, and more specifically how predictions \emph{would have changed} had relevant types been manipulated, we first construct sets of entity types for each class as approximations of what a non-expert might come up with by simple string matching per class against the 68K entity types in the type system provided in \cite{garcia-olano-etal-2021-biomedical}.  These terms for inclusion and exclusion from the sets along with the resulting type set sizes are provided in Appendix Table \ref{tab:coarse-sets}. 
We emphasize that these were easy to assemble and are coarse, noisy sets that roughly approximate entity types we would expect to be associated with each class.

Some classes such as Organism, Organism substance, Organism subdivision and Organ have sets containing the same entity types to show even quite noisy sets can be useful.  
Our intent here is not to obtain the maximum possible accuracy we can get via entity type manipulation for error cases, but rather to show the utility of this model even when paired with noisy term sets. 

After constructing coarse sets of entity types, we identify three strategies of interest for manipulating entity types during inference time:
\begin{itemize}
    \item ``Fixing'' bad entity types (i.e., minimize the weights of entity types from the incorrectly predicted class's coarse type set).
    \item ``Promoting'' good types (i.e., maximize the weights of entity types associated with the true label's type set).
    \item Using both the fix and promote strategies together.
\end{itemize}


For our experiment, we take test error cases and for each, run them through our model and either lower (``fix'') types associated with the incorrect class set, increase (``promote'') types associated with the true class set or do ``both'' to the corresponding entity type weights in the intermediate entity types layer.  
We then observe how the final class probabilities for the task are affected by the manipulation. 
Appendix Figure \ref{fig:singlemanip} shows how a single test example's class prediction distribution, derived from its original inferred types and logits, are changed by these techniques.

\renewcommand{\arraystretch}{1}
\begin{table}[t]
\centering
\small
\begin{tabular}{lc} 
\toprule
Model & Test Accuracy \\
\midrule
ItsIRL & 91.48  \\
+ Fix types & 93.91 \\
+ Promote types & 95.74 \\
+ Both fix \& promote & 95.68 \\
+ Best of 3 ``oracle" & \textbf{96.78} \\
PubMedBERT* & 96.10 \\
\bottomrule
\end{tabular}
	\caption{Entity type manipulation results using class-specific coarse type sets 
	}
	\label{tab:manip-res}
\end{table}

\paragraph{Results} In Table \ref{tab:manip-res} we report the results for our three entity manipulation techniques using coarse term sets including the best accuracy that could have been achieved amongst them for each error pattern. The model predicting \texttt{Gene} when the true class label was \texttt{Chemical} is the most common test error pattern and in Table \ref{tab:errorsfixedbody} we show the most frequent error patterns observed on the test set.  Promoting entity types of the true class improves our model results from 91.48 to 95.74, while both promoting and fixing leads to a similar 95.68. These strategies give results on par with using a dense non-interpretable PubMedBert model while using the best among them outperforms PubMedBert.  For future work, determining the best method for each error case could be done by observing performance of the techniques on a holdout set. Fixing incorrect entity types alone under performs the other techniques possibly since down weighing incorrect types alone does not necessarily push the embedding towards the correct class.  We note these automated methods require knowledge of if and in what way initial predictions may be erroneous, and our intent is to show that manipulating entity types in ItsIRL affects classification in an intuitive way which amongst other things allows them to be used with the rule based diagnostics from prior IERs.

\renewcommand{\arraystretch}{1}
\begin{table}[t]
\small
\setlength{\tabcolsep}{4pt}
\centering
\begin{tabular}{llrrrrr}
\toprule
\textcolor{blue}{True} & \textcolor{red}{Predicted} &  Errs &  T1+2 &  T1 &  T2 &  Best\% \\
\midrule             
Chemical & Gene &          65 &        64 &           48 &       59 &   98.4 \\

Cell &   Cancer &          41 &        31 &           41 &        0 &   100 \\   
                  
Cell &  Gene &          34 &        34 &           34 &        0 &    100 \\

Multi-Tis & Tissue* &          22 &         0 &            0 &        7 &   31.8 \\

Gene &  Chemical &          17 &         3 &            3 &       10 &    58.8 \\

Organ &  Tissue &          16 &        12 &           10 &       12 &    75 \\   

Cancer &    Cell &          16 &         0 &           14 &        0 &    87 \\   

Gene  &   Organism &          15 &         6 &            0 &       15 &  100 \\
                  
Cell &   Chemical &          14 &        14 &           14 &        4 &    100 \\

Amino &  Gene &          14 &        14 &           14 &       14 &  100 \\

Pathol &  Cancer &          14 &         0 &            0 &        0 &     0 \\

Organism &   Cell &          14 &         0 &            0 &        0 &   0 \\   

Organism &   Gene  &          12 &         0 &            2 &        0 &   16.7 \\

Organ &  Multi-Tissue &          10 &         0 &            1 &        0 &    10 \\

Multi-Tis &   Cancer &          10 &         0 &            0 &        0 &    0 \\

Chemical &   Amino &          10 &        10 &           10 &       10 &   100 \\

Cancer & Org. Sub. &          10 &        10 &           10 &        0 &   100 \\

Cell &  Tissue &          10 &        10 &           10 &        5 &   100 \\   

Cell &  Celu Comp* &          10 &        10 &           10 &        0 &   100 \\
\midrule
 &                 Raw Total &         592 &       292 &          296 &      169 &  361 \\ 
 &                 Percent &          100 &       49.3 &         50 &      46.8 &  61 \\ 
\bottomrule
\end{tabular}
\caption{Most frequent error patterns and manipulation results on test data for ``Promote" (T1), ``Fix" (T2) and ``Both" (T1+2) techniques. * means the term sets are equal and as ``Fix" is first applied followed by ``Promote", the ``Both" results for these cases are identical to the ``Promote" ones. }
\label{tab:errorsfixedbody}
\end{table}

In Table \ref{tab:errorsfixedbody} we show how the entity type manipulation techniques perform on each error pattern.  
Using the best technique for each error pattern allows us to correct 361 out of 592 test errors ($\sim$61\%). 
``Promoting" types is best or tied 11 out of 15 times, ``Both" gives 10 out of 15 while ``Fixing" gives 6 out of 15.  
Given the coarse type sets, all methods work poorly on the following error patterns (True Class-Predicted): Pathological Formation-Cancer, Organism-Cell, Organism-Gene, Organ-Multi-Tissue, and Multi-Tissue-Cancer.
This suggests these sets should be edited in order to better discriminate between these classes. Resolving errors is dependent on the distance between two classes and for Cell-Cancer, Cell-Gene, Cancer-Cell and Cancer-Organism subdivision, fixing incorrect types does poorly (0 errors resolved out of 101) while at the same time, promoting types from the true class does very well resolving 99 out of 101 error cases. 
We note that this process was entirely automated and having experts edit or choose better terms to form type sets associated with each class would easily improve its performance in particular with regard to error patterns where all strategies performed poorly.

\section{Related Work}
\label{section:related-work}
In this work we introduced an architecture with an encoder that uses supervision from a pre-defined static entity type system to learn an intermediate, interpretable high dimension, sparse entity type layer which is then used by a decoder network for downstream tasks.  The most similar area of work to ours is that of Concept Bottlenecks (CBs) \cite{conceptwhitening,conceptbottlenecks} which use an encoder and supervision to learn a low dimensional, dense representation for a single task.  Supervision for CBs are hand collected by experts, dense (mostly nonzero) and exist in a low dimensional space (tens to hundreds of dimensions). For the two experiments in \cite{conceptbottlenecks} 112 binary (CUB) and 10 ordinal (OAI) concepts were gathered from experts. On the other hand, IERs and our work use static, noisy entity systems gathered via weak supervision that exist in a high dimensional space (68,340 entity types) and are pre-trained for use in downstream tasks.  Due to its size compared to layers in the rest of the network, our intermediate entity type layer is not a ``bottleneck" in the usual sense of latent spaces of autoencoders, such as those from the CB literature.  

Our use of the intermediate interpretable entity layer to represent classes for global explainability is reminiscent of work for learning prototypes for images \cite{rudinprotos}, timeseries \cite{garciaolano2019protos} or text \cite{das-acl22}, however in our case constructing the prototypes of each class is done post-hoc and as such the prototypes are used for analysis rather than classification or learning. 
Additionally, our method is interpretable at the vector component level whereas the latent representations used for constructing prototypes are not. Also, our pre-trained representations are not tied to a classification task like prototypes and as such can be used for various different tasks.

Our model could be viewed as including an internal Probing task which tests a models’ ability to induce type information by measuring the accuracy of a probe \cite{Matthew_Peters_18,John_Hewitt_19_a, John_Hewitt_19_b}.  However, probing is usually a post-hoc means of revealing the information implicitly stored within internal dense output representations, whereas our model was defined and pre-trained in such a way as to explicitly provide intermediate interpretable entity type representations.   


\section{Conclusions}
In this work we proposed Intermediate Entity-based Sparse Interpretable Representation Learning (ItsIRL), an extension to the IERs architecture which provides an intermediate interpretable layer whose decoded dense representation output can be fine-tuned and leveraged for performance on downstream tasks. Empirically we show the model substantially outperforms prior IERs work on two diverse benchmark biomedical tasks.  

To demonstrate the utility of the kind of interpretability afforded by ItsIRL, we proposed a counterfactual entity type manipulation analysis which allows for modeling debugging. 
This is a fine-grained, human interaction inquiry made possible by the proposed model architecture and pre-training scheme.  Using coarse class type sets, we show this technique can allow ItsIRL to surpass performance against dense non-interpretable models.  This analysis establishes that entity type manipulation works intuitively as expected in ItsIRL, which is important for future work on methods for flagging when a predicted answer should be inspected and possibly manipulated at the entity type level.

 We finally show how combining entity types over classes on the training set to create positive and negative class prototypes can be used to explain task specific global structure and semantics learned by our model.

\section*{Ethical Considerations}

NLP models are increasingly used in biomedicine, where some applications can be quite high-stakes. 
Establishing trust in such models is therefore paramount; unfortunately, deep neural networks tend to be opaque in their operations, potentially precluding their use in certain areas of biomedicine where they might otherwise be beneficial.
This work is a step towards more transparent NLP models.

\bibliographystyle{acl_natbib}
\bibliography{acl2021}

\newpage
\clearpage
\appendix

\begin{figure*}[!t]
    \includegraphics[width=1\linewidth]{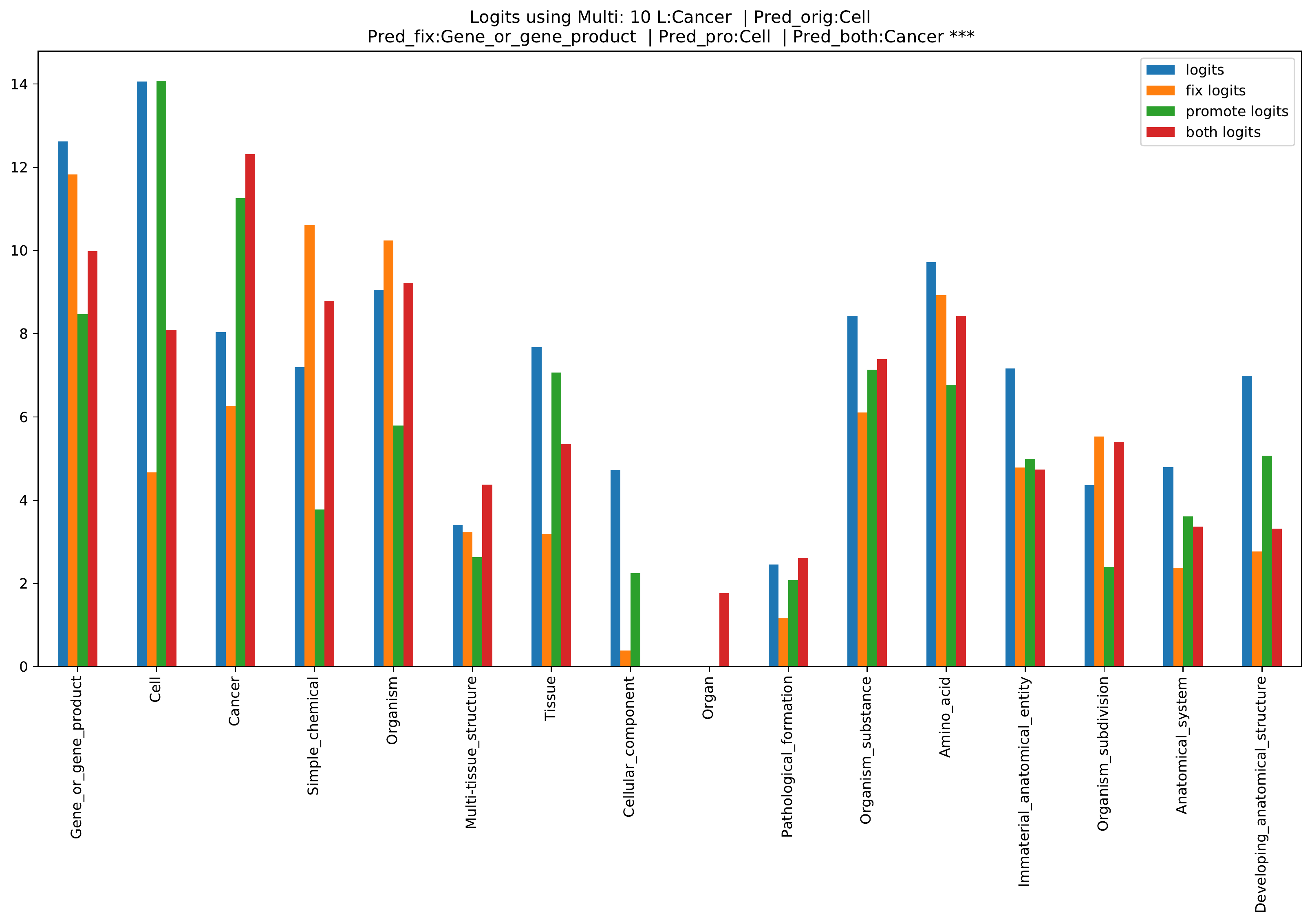}
        \caption{Class shifts using type manipulation techniques for single example}
    \label{fig:singlemanip}
\end{figure*}

\begin{figure*}[!t]
    \centering
    \includegraphics[width=.7\linewidth]{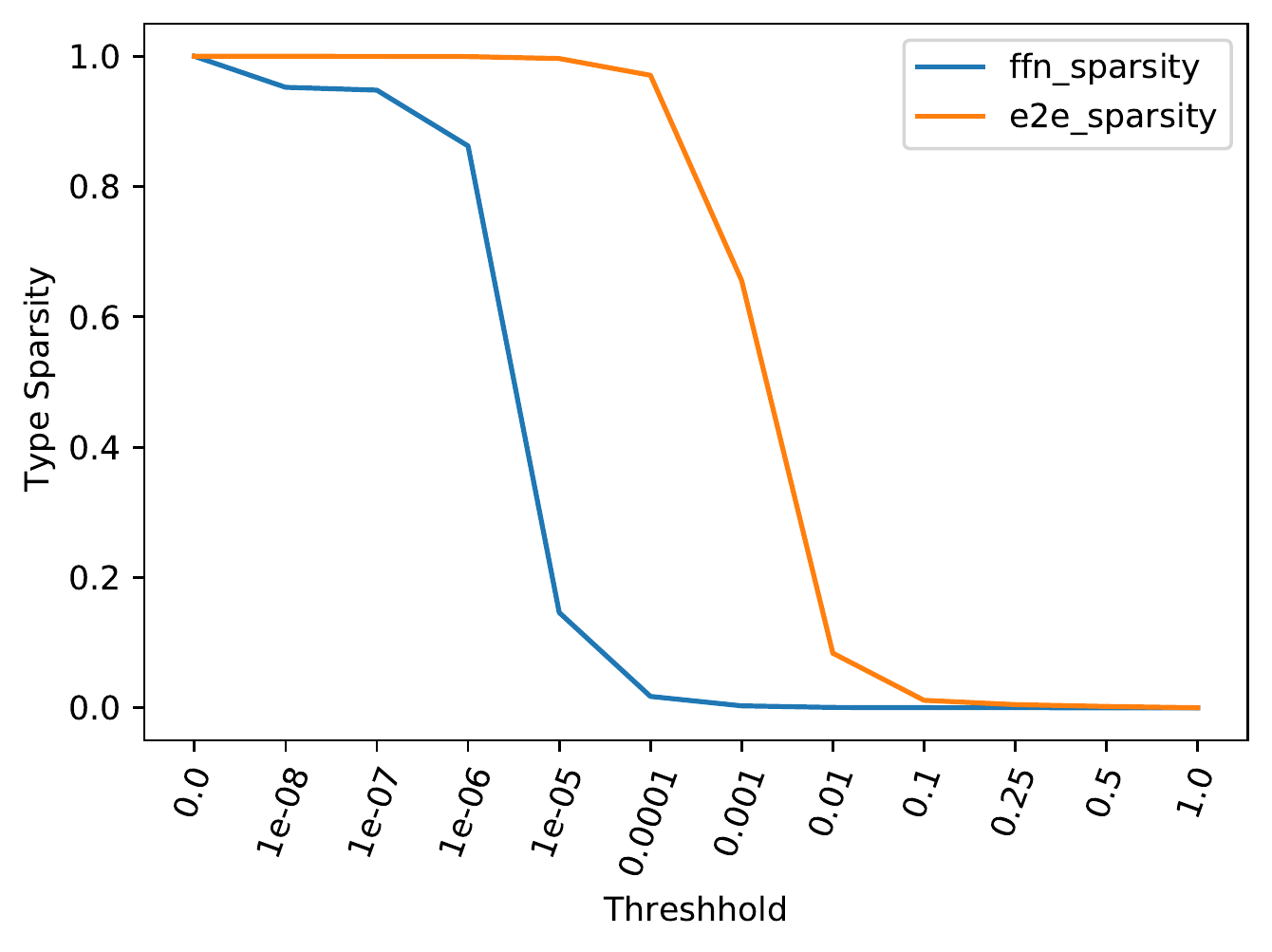}
        \caption{Entity Type Sparsity at various thresholds on BIOSSES test set}
    \label{fig:sparsity}
\end{figure*}

\newpage
\clearpage

\renewcommand{\arraystretch}{1}
\begin{table}[]
\begin{center}
\begin{tabular}{llr}
\toprule
Class &                     Term Rules Inclusion/Exclusion &  Terms in Set \\
\midrule
Cell &                                             [cell] &           357 \\
Cellular component &                                             [cell] &           357 \\
Cancer &                                 [cancer, neoplasm] &           155 \\
Gene or gene product &  \makecell[l]{[` gene', `gene ', ` genes', `genes '] \\ not in [`generation', `general']} , ' &           434 \\
Simple chemical &                             [ chemical, chemical ] &            80 \\
Organism &  \makecell[l]{[` organ', `organ ', `organism'] \\ not in [`organization']} &           172 \\
Organism substance &  \makecell[l]{[` organ', `organ ', `organism'] \\ not in [`organization']} &           172 \\
Organism subdivision &  \makecell[l]{[` organ', `organ ', `organism'] \\ not in [`organization']} &           172 \\
Organ &  \makecell[l]{[` organ', `organ ', `organism'] \\ not in [`organization']} &           172 \\
Tissue &                                 [ tissue, tissue ] &            15 \\
Multi-tissue structure &                                 [ tissue, tissue ] &            15 \\
Amino acid &                       [ amino, amino , amino acid] &            22 \\
Pathological formation &                                     [pathological] &             3 \\
Immaterial anatomical entity &             [anatomical ,  anatomical, anatomical] &            11 \\
Developing anatomical structure &             [anatomical ,  anatomical, anatomical] &            11 \\
Anatomical system &             [anatomical ,  anatomical, anatomical] &            11 \\
\bottomrule
\end{tabular}
\caption{Terms used to create coarse Class specific Entity Type sets}
\label{tab:coarse-sets}
\end{center}
\end{table}

\begin{table*}[]
\resizebox{\textwidth}{!}{%
\begin{tabular}{lllllll}
\toprule
\makecell[tl]{Gene or\\gene product} & Cell & Cancer & \makecell[tl]{Simple\\chemical} & Organism & \makecell[tl]{Multi-tissue\\structure} & Tissue  \\
\midrule
\makecell[tl]{protein\\(1.0, 5)} &              \makecell[tl]{cell\\(biology)\\(1.0, 3)} &                     \makecell[tl]{disease\\(1.0, 2)} &                  \makecell[tl]{ingredient\\(1.0, 1)} &         \makecell[tl]{taxonomy\\(biology)\\(1.0, 45)} &                      \makecell[tl]{blood\\(1.0, 47)} &         \makecell[tl]{tissue\\(biology)\\(1.0, 34)} \\

\makecell[tl]{ingredient\\(0.742, 1)} &     \makecell[tl]{elementary\\particle\\(0.346, 314)} &                  \makecell[tl]{neoplasm\\(0.897, 8)} &                     \makecell[tl]{acid\\(0.304, 18)} &  
\makecell[tl]{mammals\\described\\in 1758\\(0.943,169)} &               \makecell[tl]{angiology\\(0.843, 857)} &          \makecell[tl]{cell\\(biology)\\(0.878, 3)} \\

\makecell[tl]{human\\(0.729, 7)} &             
\makecell[tl]{human cells\\(0.201, 145)} &                 \makecell[tl]{oncology\\(0.684, 28)} &                       \makecell[tl]{rtt\\(0.301, 4)} &                  \makecell[tl]{humans\\(0.943, 187)} &            
\makecell[tl]{soft tissue\\(0.792, 3067)} &             \makecell[tl]{human\\body\\(0.814, 30)} \\

\makecell[tl]{gene\\(0.679, 6)} &   \makecell[tl]{battery\\(electricity)\\(0.192, 485)} &         \makecell[tl]{tissue\\(biology)\\(0.646, 34)} &  
\makecell[tl]{world health\\organization\\essential\\medicines\\(0.269, 25)} &      
\makecell[tl]{tool-using\\mammals\\(0.943, 186)} &                \makecell[tl]{nephron\\(0.761, 1951)} &     \makecell[tl]{connective\\tissue\\(0.385, 937)}\\

\makecell[tl]{coagulation\\(0.361, 37)} &                      \makecell[tl]{gene\\(0.184, 6)} &   \makecell[tl]{abnormality\\(behavior)\\(0.604, 56)} &        \makecell[tl]{chemical\\compound\\(0.206, 14)} &  \makecell[tl]{anatomically\\modern\\humans\\(0.943,188)} &            \makecell[tl]{blood\\vessel\\(0.682, 327)} &      \makecell[tl]{endocrine\\system\\(0.345, 482)} \\

\makecell[tl]{cell\\(biology)\\(0.353, 3)} &                   \makecell[tl]{protein\\(0.177, 5)} &                    \makecell[tl]{cancer\\(0.582, 9)} &               \makecell[tl]{measurement\\(0.19, 12)} &           \makecell[tl]{post-\\modernism\\(0.943, 177)} &               \makecell[tl]{human\\body\\(0.538, 30)} &            \makecell[tl]{epithelium\\(0.325, 144)} \\

\makecell[tl]{cell\\growth\\(0.314, 46)} &                \makecell[tl]{pancreas\\(0.167, 498)} &                 \makecell[tl]{syndrome\\(0.492, 48)} &       \makecell[tl]{calcium\\in\\biology\\(0.175, 40)} &                  \makecell[tl]{patient\\(0.863, 13)} &         \makecell[tl]{lymphatic\\system\\(0.52, 789)} &             \makecell[tl]{angiology\\(0.322, 857)} \\

\makecell[tl]{endothelium\\(0.265, 192)} &                  \makecell[tl]{system\\(0.166, 166)} &               \makecell[tl]{malignancy\\(0.467, 20)} &                  \makecell[tl]{hydroxyl\\(0.16, 76)} &       \makecell[tl]{medical\\terminology\\(0.84, 11)} &         \makecell[tl]{lymphoid\\organ\\(0.498, 1640)} &          \makecell[tl]{blood\\vessel\\(0.319, 327)}  \\

\makecell[tl]{homology\\(biology)\\(0.241, 111)} &         \makecell[tl]{carboxylic\\acid\\(0.164, 577)} &              \makecell[tl]{cell\\growth\\(0.466, 46)} &                 \makecell[tl]{glucose\\(0.142, 278)} &         \makecell[tl]{prothrombin\\time\\(0.836, 22)} &  \makecell[tl]{mononuclear\\phagocyte\\system\\(0.493, 979)} &             \makecell[tl]{histology\\(0.317, 391)} \\

\makecell[tl]{oncogene\\(0.24, 285)} &                   \makecell[tl]{ester\\(0.164, 208)} &  
\makecell[tl]{paraneoplastic\\ syndromes\\(0.458,380)} &             \makecell[tl]{methyl\\group\\(0.131, 72)} &                     \makecell[tl]{bbc\\(0.739, 180)} &                   \makecell[tl]{gland\\(0.471, 174)} &              \makecell[tl]{barcode\\(0.311, 2351)}  \\
\midrule
F1score - 96.29	& 90.71	& 92.73	& 90.24	& 94.10	& 81.65	& 74.94 \\
Support - 2520 & 1054 & 925 & 727 & 543 & 303 & 190	\\
\bottomrule
\end{tabular}
}
\caption{Top 10 Entity Types for 7 most frequent positive Prototype classes with weights and index of type.  F1 score and support for each class over test data is given in final two rows.}
\label{tab:pos8full}
\end{table*}

\begin{table*}[]
\resizebox{\textwidth}{!}{%
\begin{tabular}{lllllllll}
\toprule
\makecell[tc]{Cellular\\component} &  Organ &                 \makecell[tc]{Pathological\\formation} &                            \makecell[tc]{Organism\\substance} &                                \makecell[tc]{Amino\\acid} &                       \makecell[tc]{Immaterial\\anatomical\\entity} &                      \makecell[tc]{Organism\\subdivision} &                              \makecell[tc]{Anatomical\\system} &                    \makecell[tc]{Developing\\anatomical\\structure} \\

\midrule
\makecell[tc]{dna\\(1.0, 127)} &
\makecell[tc]{tongue\\(1.0, 158)} &
\makecell[tc]{disease\\(1.0, 2)} &
\makecell[tc]{blood\\(1.0, 47)} &
\makecell[tc]{ingredient\\(1.0, 1)} &
\makecell[tc]{cell anatomy\\(1.0, 464)} &
\makecell[tc]{anatomical\\terms of\\location\\(1.0, 373)} &
\makecell[tc]{organ\\(1.0, 138)} &
\makecell[tc]{embryology\\(1.0, 3496)} \\

\makecell[tc]{ingredient\\(0.97, 1)} &
\makecell[tc]{ecosystem\\(0.88, 268)} &
\makecell[tc]{wound\\(0.85, 2492)} &
\makecell[tc]{tetrahydro-\\gestrinone\\(0.51, 828)} &
\makecell[tc]{amino acid\\(0.97, 98)} &
\makecell[tc]{cell biology\\(0.99, 84)} &
\makecell[tc]{human body\\(0.93, 30)} &
\makecell[tc]{system\\(0.91, 166)} &
\makecell[tc]{childbirth\\(0.07, 101)} \\

\makecell[tc]{molecule\\(0.89, 82)} &
\makecell[tc]{organs\\(0.75, 321)} &
\makecell[tc]{medical\\emer-\\gencies\\(0.77, 532)} &
\makecell[tc]{nitrous\\oxide\\(0.48, 16)} &
\makecell[tc]{glucogenic\\amino acids\\(0.96, 757)} &
\makecell[tc]{cell\\(0.77, 3)} &
\makecell[tc]{leg\\(0.91, 2382)} &
\makecell[tc]{nervous\\system\\(0.72, 566)} &
\makecell[tc]{midwifery\\(0.07, 1835)} \\

\makecell[tc]{acid\\(0.89, 18)} &
\makecell[tc]{human\\body\\(0.69, 30)} &
\makecell[tc]{injury\\(0.75, 463)} &
\makecell[tc]{psychosis\\(0.48, 26)} &
\makecell[tc]{protein-\\ogenic\\amino acids\\(0.96, 657)} &
\makecell[tc]{intra-\\cellular\\(0.74, 328)} &
\makecell[tc]{limb\\(0.85, 3675)} &
\makecell[tc]{central\\nervous\\system\\(0.59, 721)} &
\makecell[tc]{health issues\\in pregnancy\\(0.07, 2873)} \\

\makecell[tc]{biotech-\\nology\\(0.89, 140)} &
\makecell[tc]{organ\\(0.64, 138)} &
\makecell[tc]{morph-\\ology\\(0.75, 137)} &
\makecell[tc]{hematology\\(0.39, 236)} &
\makecell[tc]{acid\\(0.93, 18)} &
\makecell[tc]{molecular\\biology\\(0.73, 55)} &
\makecell[tc]{tongue\\(0.71, 158)} &
\makecell[tc]{central\\african\\republic\\(0.58, 4155)} &
\makecell[tc]{health care\\(0.07, 272)} \\

\makecell[tc]{polymer\\(0.89, 1204)} & 
\makecell[tc]{articles\\containing\\video clips\\(0.54, 19)} &
\makecell[tc]{injuries\\(0.75, 3237)} &
\makecell[tc]{ingredient\\(0.32, 1)} &
\makecell[tc]{calcium in\\biology\\(0.90, 40)} &
\makecell[tc]{middle east\\(0.28, 1229)} & 
\makecell[tc]{lower limb\\anatomy\\(0.67, 8420)} &
\makecell[tc]{chemical\\structure\\(0.57, 1315)} &
\makecell[tc]{fetus\\(0.07, 1172)} \\

\makecell[tc]{helices\\(0.87, 2487)} &
\makecell[tc]{human\\anatomy\\by organ\\(0.44, 1430)} &
\makecell[tc]{acute pain\\(0.75, 923)} &
\makecell[tc]{articles\\containing\\video clips\\(0.29,19)} &
\makecell[tc]{measure-\\ment\\(0.83, 12)} &
\makecell[tc]{route of\\admin-\\istration\\(0.24, 209)} &
\makecell[tc]{anatomy\\(0.63, 287)} &
\makecell[tc]{cerebro-\\spinal\\fluid\\(0.56, 2756)} &
\makecell[tc]{obstetrical\\procedures\\(0.0, 146)} \\

\makecell[tc]{nucleic acids\\(0.89, 1426)} &
\makecell[tc]{gland\\(0.43, 174)} &
\makecell[tc]{first aid\\(0.74, 5588)} &
\makecell[tc]{cell\\anatomy\\(0.27, 464)} &
\makecell[tc]{amine\\(0.67, 61)} &
\makecell[tc]{abdomen\\(0.24, 503)} &
\makecell[tc]{animal\\locomotion\\(0.62, 672)} &
\makecell[tc]{musical\\quintets\\(0.5, 1926)} &
\makecell[tc]{blood cells\\(0.0, 2195)} \\

\makecell[tc]{cell\\(0.83, 3)} &
\makecell[tc]{digestion\\(0.39, 607)} &
\makecell[tc]{physical\\therapy\\(0.73, 1765)} &
\makecell[tc]{tissues\\(0.27, 791)} &
\makecell[tc]{isomer\\(0.48, 800)} &
\makecell[tc]{drug\\(0.19, 24)} &
\makecell[tc]{foot\\(0.59, 5959)} &
\makecell[tc]{radiophar\\macology\\(0.49, 3611)} &
\makecell[tc]{developmental\\biology\\(0.0, 352)} \\

\makecell[tc]{cell\\membrane\\(0.58, 288)} &
\makecell[tc]{tissue\\(0.38, 34)} &
\makecell[tc]{tongue\\(0.37, 158)} &
\makecell[tc]{body fluids\\(0.19, 617)} &
\makecell[tc]{ketogenic\\amino acids\\(0.42, 1974)} &
\makecell[tc]{pharma-\\ceutical\\drug\\(0.18, 17)} &
\makecell[tc]{animal\\(0.59, 273)} &
\makecell[tc]{earache\\records\\(0.48, 5219)} &
\makecell[tc]{transformation\\(genetics)\\(0.0, 752)} \\
\bottomrule
\end{tabular}
}
\caption{Top 10 Entity Types for remaining 9 positive Prototype classes with weights and index of type.  F1 score and support for each class over test data is given in final two rows.}
\label{tab:pos8fullb}
\end{table*}

\begin{table*}[]
\resizebox{\textwidth}{!}{%
\begin{tabular}{llllllll}
\toprule

\textcolor{blue}{Truth} & Cell & Chemical & Cell & Organism & Tissue & Gene & Cancer \\
\textcolor{red}{Pred} & Cancer & Gene & Gene & Gene & Multi-tissue & Chemical & Cell \\
\midrule
1 &
\makecell[tl]{cancer\\(1.0, 9)} &                  \makecell[tl]{ingredient\\(1.0, 1)} &                       \makecell[tl]{gene\\(1.0, 6)} &                        \makecell[tl]{gene\\(1.0, 6)} &             
\makecell[tl]{histology\\(1.0, 391)} &                  \makecell[tl]{ingredient\\(1.0, 1)} &       \makecell[tl]{cell\\(biology)\\(1.0, 3)} \\ 
2 &
\makecell[tl]{disease\\(0.87, 2)} &                   \makecell[tl]{protein\\(0.61, 5)} &                  \makecell[tl]{protein\\(0.65, 5)} &                   \makecell[tl]{protein\\(0.93, 5)} &                
\makecell[tl]{blood\\(0.96, 47)} &                     \makecell[tl]{acid\\(0.58, 18)} &           
\makecell[tl]{neoplasm\\(0.41, 8)} \\ 
3 &
\makecell[tl]{neoplasm\\(0.73, 8)} &  \makecell[tl]{receptor\\(biochemistry)\\(0.53, 52)} &                    \makecell[tl]{human\\(0.50, 7)} &                     \makecell[tl]{human\\(0.65, 7)} &        
\makecell[tl]{blood\\vessel\\(0.96, 327)} &         \makecell[tl]{chemical\\compound\\(0.53, 14)} &            \makecell[tl]{disease\\(0.38, 2)} \\ 
4 &
\makecell[tl]{malignancy\\(0.66, 20)} &                      \makecell[tl]{gene\\(0.49, 6)} &                  
\makecell[tl]{allele\\(0.34, 71)} &                   \makecell[tl]{allele\\(0.43, 71)} &           
\makecell[tl]{angiology\\(0.92, 857)} &   \makecell[tl]{derivative\\(chemistry)\\(0.42, 58)} &           \makecell[tl]{t\\cell\\(0.36, 429)} \\
5 &
\makecell[tl]{rtt\\(0.55, 4)} &                     
\makecell[tl]{human\\(0.41, 7)} &               \makecell[tl]{ingredient\\(0.28, 1)} &                \makecell[tl]{apoptosis\\(0.37, 87)} &             \makecell[tl]{nephron\\(0.74, 1951)} &                   \makecell[tl]{protein\\(0.34, 5)} &       
\makecell[tl]{lymphocyte\\(0.35, 112)} \\
6 &
\makecell[tl]{oncology\\(0.46, 28)} &                    \makecell[tl]{enzyme\\(0.34, 29)} &  \makecell[tl]{receptor\\(biochemistry)\\(0.25, 52)} &               \makecell[tl]{wild\\type\\(0.35, 159)} &  \makecell[tl]{circulatory\\system\\(0.64, 664)} &                  \makecell[tl]{purine\\(0.32, 781)} &             
\makecell[tl]{cancer\\(0.25, 9)} \\ 
7 &
\makecell[tl]{squamous-\\cell\\carcinoma\\(0.37, 163)} &                    \makecell[tl]{blood\\(0.29, 47)} &  
\makecell[tl]{transcription\\factors\\(0.25, 219)} &                 \makecell[tl]{ingredient\\(0.34, 1)} &              \makecell[tl]{tongue\\(0.58, 158)} &        \makecell[tl]{deciduous\\teeth\\(0.28, 3292)} &     \makecell[tl]{lymphoblast\\(0.25, 1200)} \\
8 &
\makecell[tl]{tissue\\(biology)\\(0.35, 34)} &     \makecell[tl]{receptor\\antagonist\\(0.28, 922)} &              \makecell[tl]{coagulation\\(0.23, 37)} &           \makecell[tl]{fas\\receptor\\(0.33, 5278)} &                \makecell[tl]{heart\\(0.54, 353)} &             \makecell[tl]{cell\\(biology)\\(0.27, 3)} &           \makecell[tl]{thymus\\(0.23, 506)} \\
9 &
\makecell[tl]{cell\\(biology)\\(0.31, 3)} &         \makecell[tl]{enzyme\\inhibitor\\(0.28, 41)} &             \makecell[tl]{cell\\growth\\(0.23, 46)} &  \makecell[tl]{tumor\\necrosis\\factor\\alpha\\(0.30, 604)}  &              \makecell[tl]{kidney\\(0.52, 430)} &                  \makecell[tl]{tooth\\(0.27, 2205)} &              
\makecell[tl]{human\\(0.22, 7)} \\ 
10 &
\makecell[tl]{infectious\\causes of\\cancer\\(0.30, 73)} &                  \makecell[tl]{antigen\\(0.27, 64)} &                    \makecell[tl]{dna\\(0.21, 127)} &                  \makecell[tl]{antigen\\(0.23, 64)} &        
\makecell[tl]{soft tissue\\(0.51, 3067)} &  \makecell[tl]{receptor\\(biochemistry)\\(0.27, 52)} &  \makecell[tl]{precursor\\cell\\(0.17, 2220)} \\
\bottomrule
\end{tabular}
}
\caption{Top 10 Entity Types for 7 most frequent negative Prototypes}
\label{tab:neg7}
\end{table*}

\end{document}